\documentclass[letterpaper, 10 pt, conference]{ieeeconf}  

\IEEEoverridecommandlockouts                              
\usepackage{amssymb}
\usepackage{multicol}
\usepackage{multirow}
\usepackage{makecell}
\usepackage[bookmarks=true]{hyperref}
\usepackage{graphicx}
\usepackage{subfigure}
\usepackage{booktabs}
\usepackage[table]{xcolor}
\usepackage{array}
\usepackage{amsmath}
\usepackage{float}
\usepackage{caption}
\usepackage{comment}
\usepackage{wrapfig}

\title{\LARGE \bf
Learning in ImaginationLand:\\Omnidirectional Policies through 3D Generative Models (OP-Gen)
}

\author {Yifei Ren$^{1}$ and Edward Johns$^{1}$}

\begin{document}

\twocolumn[{
\renewcommand\twocolumn[1][]{#1}
\maketitle
\thispagestyle{empty}
\pagestyle{empty}
\vspace*{-1\baselineskip}
\begin{center}
    \centering
    \captionsetup{type=figure}
    \includegraphics[width=\linewidth]{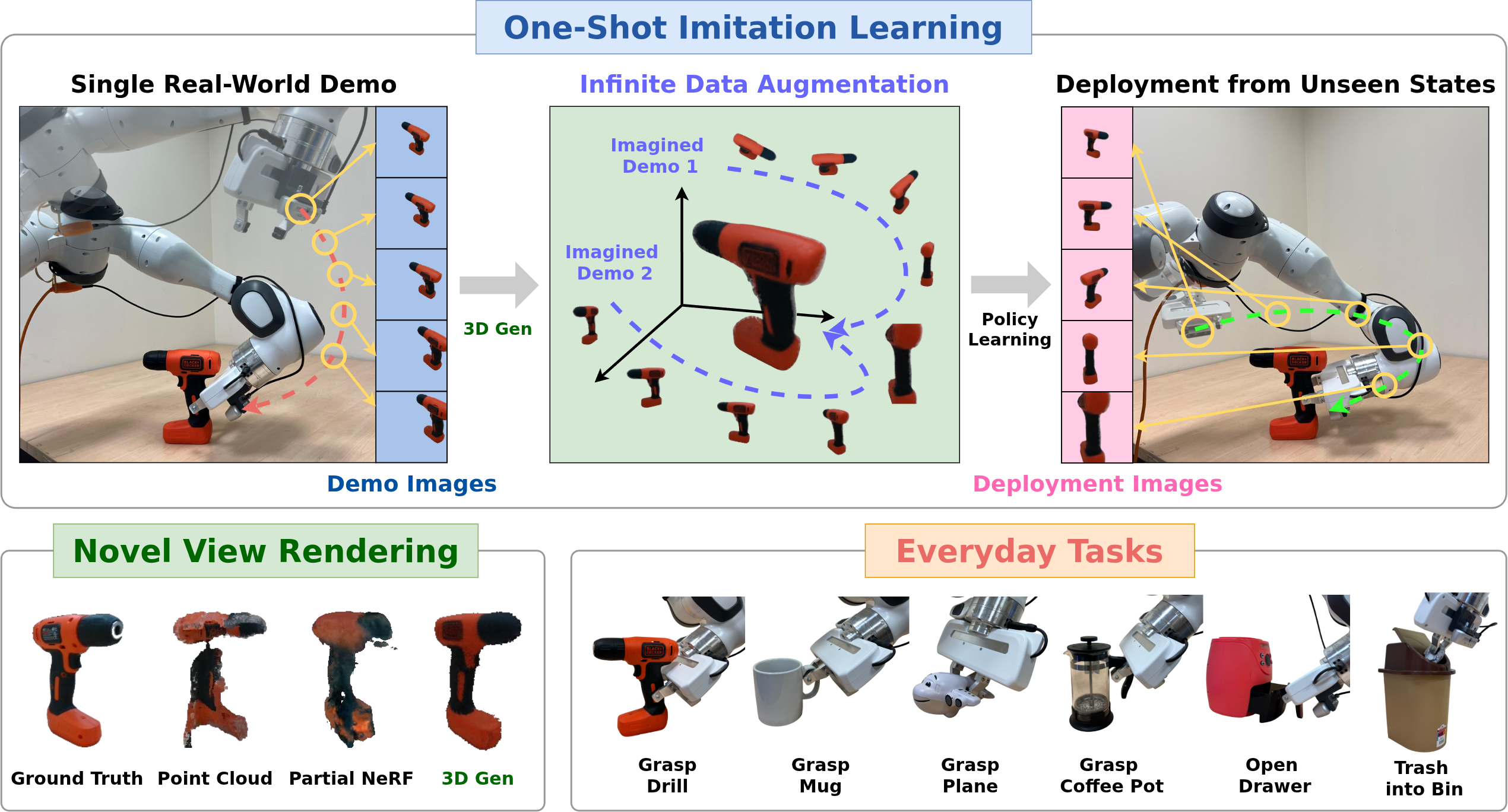}
    \caption{We utilise 3D generative models for data augmentation to enable one-shot imitation learning from a single demonstration, with end-to-end policies from a wrist-mounted RGB camera. Given a single real-world demonstration, a full 3D model of the object is generated through the generative model, which allows for 3D augmentation of an infinite number of ``imagined" demonstrations via novel view rendering. The augmented dataset is then used to train an ``omnidirectional" policy, enabling tasks to be executed from any initial state, even if that state is significantly different from any in the real-world demonstration. Leveraging the generalisation and high-quality novel view rendering capabilities of 3D generative models, our method can be applied to multiple everyday tasks.}
    \label{fig:fig1}
\end{center}
}]

%%%%%%%%%%%%%%%%%%%%%%%%%%%%%%%%%%%%%%%%%%%%%%%%%%%%%%%%%%%%%%%%%%%%%%%%%%%%%%%%
\begin{abstract}

Recent 3D generative models, which are capable of generating full object shapes from just a few images, now open up new opportunities in robotics. In this work, we show that 3D generative models can be used to augment a dataset from a single real-world demonstration, after which an omnidirectional policy can be learned within this imagined dataset. We found that this enables a robot to perform a task when initialised from states very far from those observed during the demonstration, including starting from the opposite side of the object relative to the real-world demonstration, significantly reducing the number of demonstrations required for policy learning. Through several real-world experiments across tasks such as grasping objects, opening a drawer, and placing trash into a bin, we study these omnidirectional policies both qualitatively and quantitatively by investigating the effect of various design choices on policy behaviour, and we show superior performance to recent baselines which use alternative methods for data augmentation. Videos of our experiments can be found on our webpage: \textcolor{blue}{\href{https://sites.google.com/view/op-gen}{https://sites.google.com/view/op-gen}.}

\end{abstract}

\let\thefootnote\relax\footnotetext{$^{1}$ The Robot Learning Lab at Imperial College London.}

%%%%%%%%%%%%%%%%%%%%%%%%%%%%%%%%%%%%%%%%%%%%%%%%%%%%%%%%%%%%%%%%%%%%%%%%%%%%%%%%
\section{INTRODUCTION}

Demonstrations are a powerful source of data for policy learning, and in recent years, this has led to impressive examples of reactive and dexterous policies \cite{black2024pi_0, zhao2024aloha}. However, typical behavioural cloning methods would likely require hundreds of demonstrations to learn a policy for the tasks in Figure \ref{fig:fig1}. This is because to learn robust omnidirectional policies (OP) --- which can succeed from any initial state --- the demonstrations must densely cover a very large distribution of states.

%%%%%%%%%%%%% Original Version %%%%%%%%%%%%
In this work, we ask whether these omnidirectional policies can instead be learned from only a single demonstration with a wrist-mounted RGB camera, by bringing in recent 3D generative models \cite{kong2024eschernet}. Having been trained on huge datasets of objects, these models enable images to be imagined from any viewpoint of a new object, when given only a small number of images of that object. For example, they can produce the “back” of an object when only its “front” is observed. Given real images observed during a single demonstration, we therefore propose that images from novel viewpoints can be imagined and automatically labelled with actions. A policy trained on this ``imagined" dataset would then be in-distribution for any view of the object, even if that view is from the opposite side of the object from where the demonstration was collected.

%%%%%%%%%%%%% Original Version %%%%%%%%%%%%
This can be seen as a form of dataset augmentation. Prior works such as DemoGen \cite{xue2025demogen} reconstruct partial point clouds from demonstrations for this purpose, but conventional point cloud methods cannot recover unseen geometry accurately, limiting generalisation when those regions are relevant at test time. Fortunately, with the arrival of 3D generative AI, we can now show how full 3D augmentation enables generalisation to viewpoints that were previously entirely unobserved.

%%%%%%%%%%%%% New Version %%%%%%%%%%%%
Building on this insight, we introduce \textbf{OP-Gen}, a framework for learning omnidirectional policies from a single demonstration using 3D generative models. Following \cite{xue2025demogen}, we decompose the demonstration into a \textit{motion} phase that reaches a suitable preconditioned state and a \textit{skill} phase that executes the task behaviour. There are multiple ways to obtain the \textit{skill} component. For simplicity, we adopt a replay-based implementation, although it can also be instantiated using \textit{skills} learned via imitation learning. From a few wrist-camera images, we generate a full 3D object model and collision-free end-effector trajectories that guide the robot from novel initial poses to the start of the demonstrated \textit{skill}. Rendering synthetic images along these trajectories yields a large ``imagined" dataset of image-action pairs used to train a diffusion policy \cite{chi2023diffusion}, producing robustness across a wide range of initial poses where the object remains visible.

To the best of our knowledge, this is the first method for 6-DoF policy learning that uses 3D generative models for single-demo augmentation from a wrist-mounted RGB camera. Since current 3D generative models are typically designed for single-object generation, in our real-world experiments, we focus on single-object tasks such as grasping objects, opening a drawer, and placing trash into a bin. However, as generative models improve, this method could be extended to full scenes. We show that our omnidirectional policies outperform alternative methods for 3D augmentation, such as rendering images from a point cloud or a NeRF reconstruction. Our ablation studies then explore how the quality of the generated 3D model affects task performance, and we find what the optimal strategy is to generate these imagined datasets when given a single demonstration.

%===============================================================================
\vspace{-5pt}
\section{Related Work}
    \label{sec:related work}
    \textbf{Efficient Imitation Learning.}
    In robotics, real-world data is sparse and expensive to collect. This makes data-efficient imitation learning crucial, yet current approaches typically require around a hundred demonstrations to train robust policies \cite{chi2023diffusion, bharadhwaj2024roboagent, jang2022bc, zhao2023learning,wang2025one}. Replay-based approaches mitigate this by positioning the robot at a ``bottleneck" pose relative to the target object before replaying the demonstrated interactions \cite{johns2021coarse, valassakis2022demonstrate, vitiello2023one, wen2022you, di2024effectiveness}, which enables one-shot imitation learning, even on contact-rich tasks \cite{papagiannis2024miles}. Novel in-context learning frameworks further enhance efficiency \cite{dipalo2024kat, vosylius2024instant}, and recent works even push the boundaries toward zero-shot manipulation \cite{kapelyukh2023dall, dream2real, kwon2024language}.
    
    \textbf{2D Data Augmentation.}
    Data augmentation has traditionally supported 2D domain randomisation through image cropping, flipping, and lighting variations to improve policy robustness  \cite{laskin2020reinforcement, li2021domain, yarats2021image}. Additionally, works such as \cite{ho2021retinagan, rao2020rl, sadeghi2017sim2real} focus on augmenting data to reduce the sim-to-real gap. ROSIE \cite{yu2023scaling} extends this by modifying backgrounds and introducing distractors, leveraging text-to-image models \cite{ramesh2021zero, ramesh2022hierarchical, rombach2022high, saharia2022photorealistic}. However, none of these techniques enable generalisation to unseen object parts. Our method seamlessly integrates with these approaches, enabling a more comprehensive form of augmentation. 
    
    \textbf{3D Data Augmentation.} Given that most robotic tasks operate in 6-DoF spaces, 3D data augmentation is highly beneficial. SPARTN \cite{zhou2023nerf} employs NeRFs \cite{mildenhall2021nerf} to perturb end-effector poses for data augmentation but is limited to nearby views. In contrast, our method uses 3D generative models to fully generate the target object, enabling arbitrary augmented trajectories. While \cite{zhu2024nerf} achieves object-level generalisation with NeRFs and \cite{zhang2024diffusion} uses diffusion models for future observation prediction under dynamic interaction, both lack coverage of unseen regions. Simulation-based methods such as MimicGen \cite{mandlekar2023mimicgen} and its variation \cite{garrett2024skillmimicgen} generate large datasets from few demonstrations but suffer from sim-to-real gaps, while geometry-aware sim-to-real approaches \cite{lin2025constraint} require complete object models. DemoGen \cite{xue2025demogen} augments real demonstrations using partial point clouds, but struggles when test states expose unseen object regions. Other real-to-sim transfer methods \cite{torne2024reconciling, kerr2024robot, han2025re, yang2025novel, barcellona2024dream} similarly depend on full geometry and dynamics, which are difficult to obtain from a single demonstration. Recent 3D generation advances have motivated related exploration \cite{nolte2025single}; for example, \cite{tian2024view} performs zero-shot novel view synthesis on observations but does not augment actions. RobotTwin \cite{mu2024robotwin} reconstructs full object models via 3D generation, yet relies on GPT-4V-generated expert code \cite{2023GPT4VisionSC} rather than human demonstrations. In contrast, our method leverages 3D generative models to synthesise full object geometry and augment both observations and executable action trajectories from a single real-world demonstration.

    \begin{figure*}[!htbp]
        \centering
        \includegraphics[width=\linewidth]{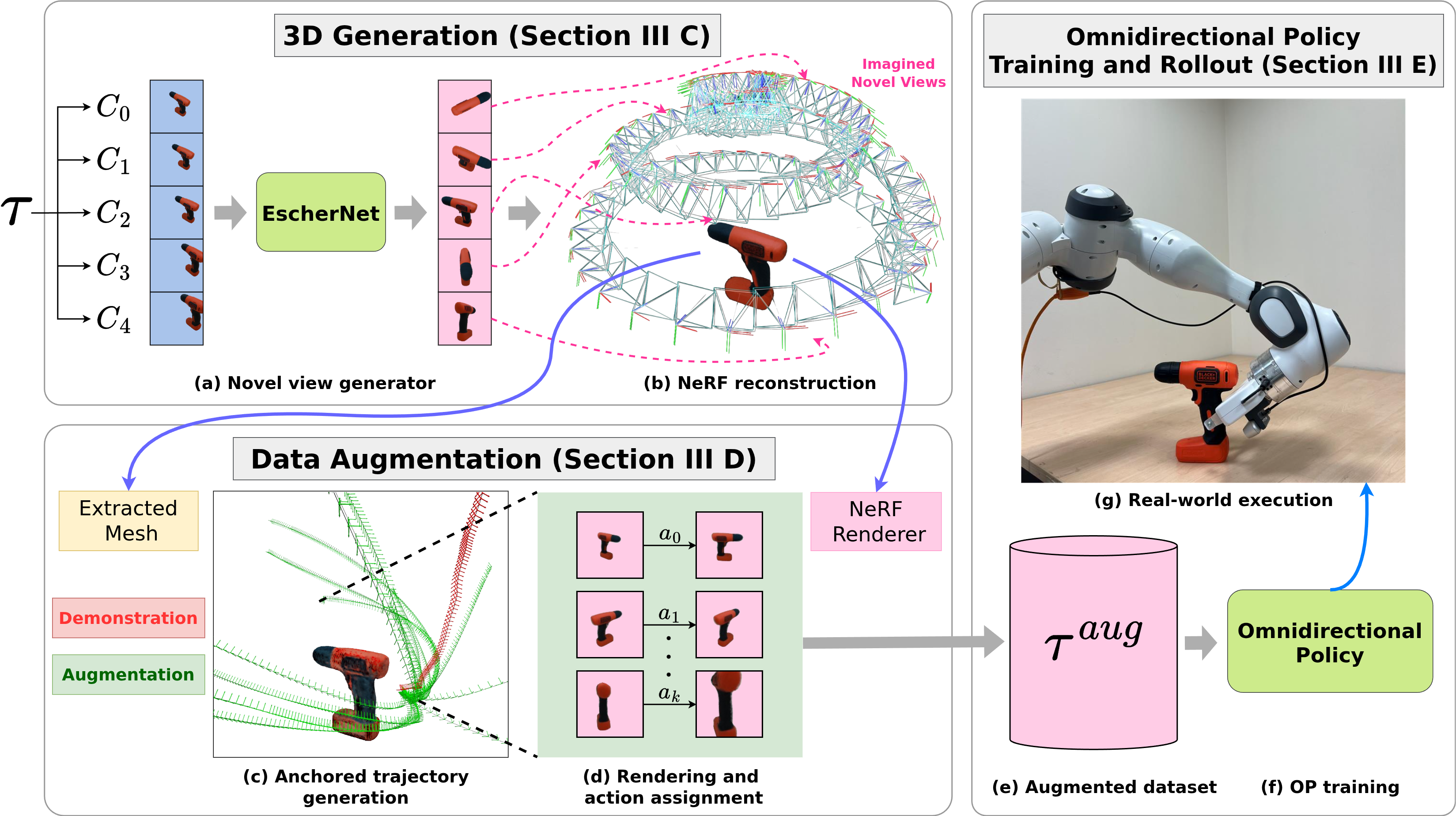}
        \caption{The OP-Gen begins with a single demonstration  $\tau$, from which posed images are sampled and fed into EscherNet \cite{kong2024eschernet} for novel view synthesis (a). The resulting multi-view images are used to construct a NeRF \cite{muller2022instant} for efficient rendering (b). The extracted 3D mesh of the target object enables our anchored trajectory generation module to create novel trajectories (c). Then we render new observations via the pre-built NeRF and assign corresponding actions (d). These are aggregated into an augmented dataset (e), used to train a diffusion policy \cite{chi2023diffusion} (f), which is then deployed in real-world rollouts (g).}
        \label{fig:pipeline}
        \vspace{-15pt}
    \end{figure*}

    \textbf{3D vision for robotics.} 
     3D vision is increasingly used in robotics due to the natural compatibility with robotic systems. \cite{mildenhall2021nerf, muller2022instant, kerbl20233d} provide efficient, photorealistic novel view synthesis, which enables various downstream applications \cite{dream2real, barcellona2024dream, abou2024physically, ji2024graspsplats}. Hoever, these approaches require dense view coverage, limiting their usage in sparse or low-data scenarios common in robotics. Recent 3D generative models \cite{liu2023zero, long2024wonder3d, voleti2025sv3d} address this gap by enabling plausible view synthesis and object generation from minimal input, but none of them ensure up-to-scale alignment to the real world. To address this, EscherNet \cite{kong2024eschernet} novelly encodes camera poses with RGB images, achieving any-to-any view synthesis while maintaining scale awareness.     

%=====================================================
%=====================================================
\section{OP-Gen: Omnidirectional Policies through 3D Generative Models}
\label{sec:method}

    \subsection{Overview}
    From a single demonstration, the target object is only partially observed, making it challenging to perform the task from unseen views. OP-Gen addresses this by using 3D generative models to generate the object and synthesise consistent renderings from unseen viewpoints. These are labeled with actions that guide the robot back to and along the demonstration, enabling training of an omnidirectional policy, as shown in Fig.~\ref{fig:pipeline}.

    \subsection{Single Demonstration}
    We consider a one-shot, RGB-only imitation learning setting, in which a single demonstration $\tau$ is provided for each task. Each demonstration comprises a sequence of observation-action tuples $\{I_t, C_t, a_t\}$, where $I_t$ denotes a $64\times64$ RGB image captured from a wrist-mounted camera, $C_t$ represents the corresponding 6-DoF camera pose, and $a_t$ indicates the 7-DoF EEF action. The EEF action consists of a 6-DoF transformation to the next pose relative to the EEF frame and a 1-DoF gripper state (open or close). All pose representations are expressed in angle-axis format. Following the conventions of \cite{xue2025demogen, johns2021coarse, vitiello2023one, wen2022you, mandlekar2023mimicgen}, each demonstration is divided into \textit{motion} and \textit{skill} phases, with ``bottleneck" points marking the transitions, which are easily identified by checking if the distance between the EEF and the centre of the target object falls within a threshold. During data augmentation, the \textit{skill} phases are directly transformed as complete segments, whereas the \textit{motion} phases are re-planned using motion planning to bridge consecutive \textit{skill} phases, consistent with \cite{xue2025demogen, zhou2023nerf}. Furthermore, given that current 3D generation techniques primarily address object-level rather than scene-level synthesis, we perform target object segmentation during both training and testing phases using XMem \cite{cheng2022xmem}, aligning with state-of-the-art practices.

    \subsection{3D Generation}
    From single-view or few-views inputs, we can synthesise novel views through 3D generative models, which is shown in Fig. \ref{fig:pipeline}(a). We use EscherNet \cite{kong2024eschernet} as our novel view generator $\mathcal{G}$ since it can generate up-to-scale 3D consistent novel views. Specifically, for each demonstration $\tau$, we sample 5 images $I_S=\{I_k\}_{k=1}^5$ and their corresponding camera poses $C_S=\{C_k\}_{k=1}^5$ as inputs to EscherNet, and then query renderings from 100 novel views $I_Q=\{\mathcal{G}(I_S, C_S, C_q)\}_{q=1}^{100}$ within a minute, where $C_q$ is the queried pose along an Archimedean spiral centered at the object as shown in Fig. \ref{fig:pipeline}(b). While EscherNet can render images directly, querying it repeatedly during data augmentation is computationally expensive and time consuming. Therefore, we build a NeRF based on the rendered images $I_Q$ using Instant-NGP \cite{muller2022instant} for faster rendering, and extract the mesh of the object $\mathcal{M}_O$ from it for subsequent collision checking.
    
    \subsection{Data Augmentation}
\begin{figure}[htbp]
        \centering
        \includegraphics[width=\linewidth]{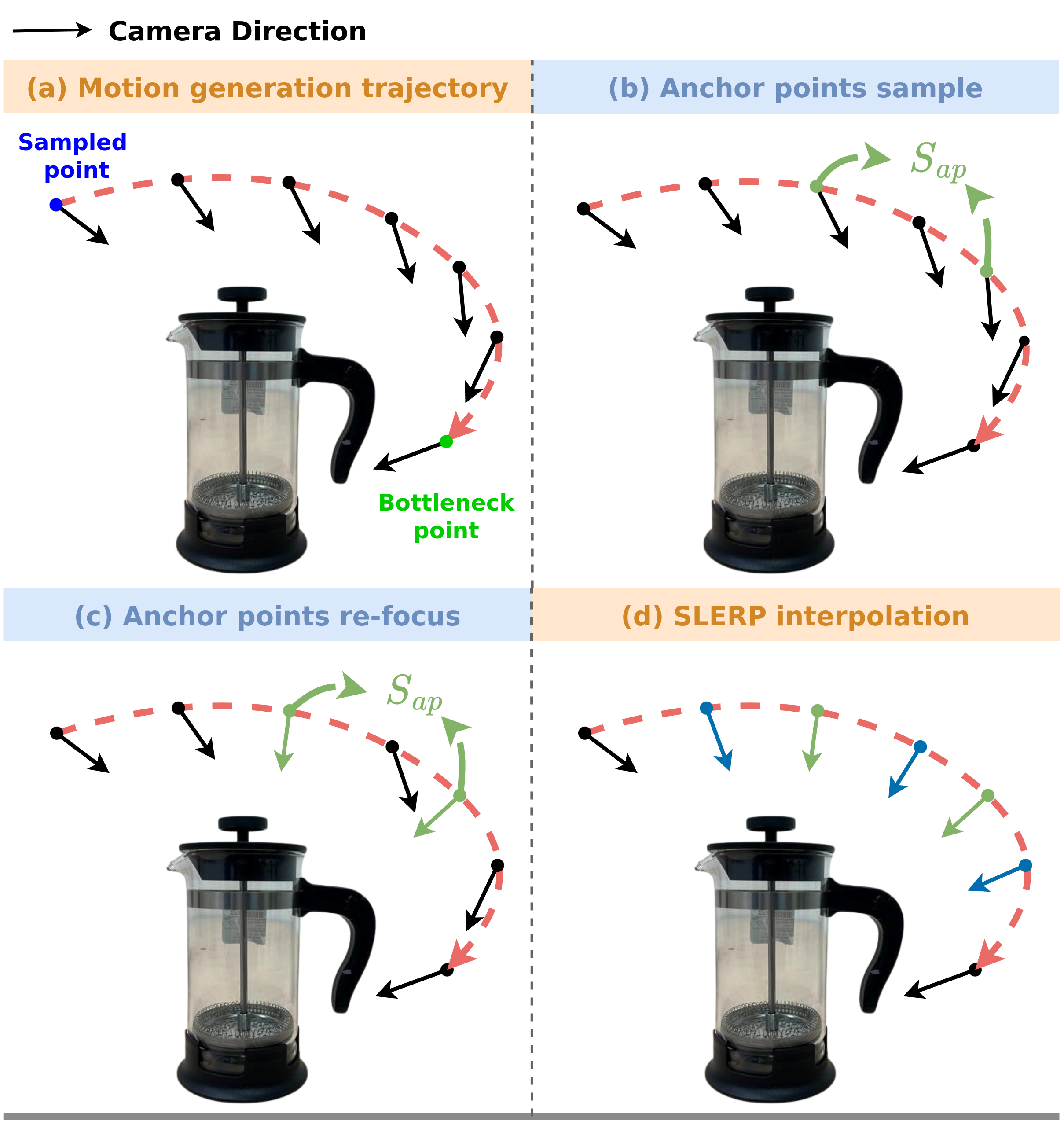}
        \caption{An illustration of how anchored trajectory generation works. (a) shows the generated collision-free trajectory, and the black arrows are the look-at directions of the cameras. Note we always fix the poses of the sampled starting point and the last bottleneck point afterward; (b) shows the evenly sampled anchor points; (c) shows the re-focus process; (d) shows the final SLERP interpolation.}
        \label{fig:anchored_traj_gen}
        \vspace{-12pt}
    \end{figure}

    By leveraging the NeRF model, we are able to render images from arbitrary sampled viewpoints. Moreover, given access to the mesh of the target object, we generate complete, collision-free trajectories from any sampled viewpoint back to the demonstration trajectory, rather than subtly perturbing demonstration actions as in \cite{zhou2023nerf}. However, in the absence of explicit constraints to ensure that the camera remains focused on the target object, the rendered images may include irrelevant or blank regions, potentially degrading policy performance. To address this, we propose an Anchored Trajectory Generation (ATG) pipeline, as illustrated in Fig.~\ref{fig:anchored_traj_gen}.
    
    \textbf{Anchored Trajectory Generation.} 
    First, as shown in Fig.\ref{fig:anchored_traj_gen}(a), an EEF trajectory $\{E_i\}_{i=1}^k$ is generated from a randomly sampled pose to the predefined bottleneck pose using CuRobo \cite{sundaralingam2023curobo}, a parallelised, collision-free motion planner, where each $E_i$ represents a 6-DoF EEF pose.
    Subsequently, anchor points are uniformly sampled along this trajectory, as denoted by $S_{ap}$ in Fig.\ref{fig:anchored_traj_gen}(b). For each anchor point, we fix its position while adjusting its orientation to ensure that the camera remains focused on the target object. To enhance generalisation, small perturbations are added to the orientation, resulting in corrected poses $E_i^{\text{correct}}$. We refer to this adjustment process as re-focus, illustrated in Fig.\ref{fig:anchored_traj_gen}(c).
    Finally, to ensure smooth rotational transitions between consecutive anchor points, the start and end points, we apply spherical linear interpolation (SLERP), yielding the augmented anchored trajectory $\{E_i^{aug}\}_{i=1}^k$, depicted in Fig.\ref{fig:anchored_traj_gen}(d).
    
    \textbf{Rendering and Action Assignment.} After acquiring the anchored trajectories, the corresponding augmented camera poses $\{C_i^{aug}\}_{i=1}^k$ are computed using the pre-calibrated camera-to-EEF transformation. These new camera poses are then used to render images $\{I_i^{aug}\}_{i=1}^k$ as novel observations via the pre-built NeRF model. Meanwhile, for each EEF pose along the trajectory, we compute the relative transformation to the subsequent pose $E_i^{-1}E_{i+1}$, and integrate it with the gripper state to form the new action set $\{a_i^{aug}\}_{i=1}^k$. The gripper state is assigned a value of 1 (close) only when the pose is within 1 cm of the bottleneck. Since fewer than 1\% of augmented poses are near the bottleneck, we oversample poses within 1 cm of it to ensure balanced data distribution. Finally, by associating rendered images with assigned actions across all generated anchored trajectories, we construct the augmented dataset $\mathcal{D}_{aug} = \{I_i^{aug}, C_i^{aug}, a_i^{aug} \}_{i=1}^K$, where $K=k_1 + k_2 + ... + k_n$ denotes the total number of generated data points. In practice, we set $K = 20000$.

    \subsection{Training and Policy Rollout}
    After obtaining the augmented dataset, we train a behavioural cloning (BC) policy to mimic actions in it. The objective is to optimise a policy $\pi_\theta$ that closely fits the augmented data. Specifically, we adopt diffusion policy \cite{chi2023diffusion} as the backbone for our approach, thus the training loss is formulated as
    \begin{equation}
        \mathcal{L}=MSE(\epsilon^k, \epsilon_\theta(I_t, a_t+\epsilon^k, k)),
        \label{eq:train}
    \end{equation}
    where $\epsilon_\theta$ represents the diffusion model used to predict the added noise at each step $k$. During testing, we predict the relative EEF pose by iteratively denoising randomly initialised actions using the DDIM algorithm \cite{song2020denoising}. Specifically, we use ResNet18 \cite{he2016deep} as our visual backbone and we set the observation horizon to 1 and the action horizon to 4. Also, we set training diffusion steps to 100 and inference diffusion steps to 16. After 100000 global training steps, we select the model with the lowest validation loss as the final policy.

%===============================================================================
\section{Experiments}
\label{sec:experiments}

    \subsection{Experimental Setup}
    In this section, we evaluate OP-Gen in a one-shot imitation learning setting. Specifically, we conduct experiments on 6 different real-world tasks: \textit{drill}, \textit{mug}, \textit{plane} and \textit{coffee pot} grasping, placing trash into a \textit{trash bin}, and opening the air fryer \textit{drawer}, as shown in Figure \ref{fig:fig1}. For each task, we collect only one demonstration trajectory of approximately 10 seconds, including an RGB sequence with the corresponding EEF poses and gripper states at $20Hz$. Then we augment the original dataset to a total of 20000 data points and train our omnidirectional policy on it. Lastly, we test the rollout policies in both simulation and the real world at $20Hz$. \\
    \textbf{Simulation.} To quantitatively measure the policy performance, we create a visual simulator for each task using Instant-NGP \cite{muller2022instant}. In detail, apart from the single demonstration, we scan the whole object from a hemispherical camera trajectory which contains about 600 posed RGB images, and build a full NeRF of the scene based on them, serving as the visual simulator. To define task success, we measure the remaining distance to the bottleneck point when the policy commands to execute the \textit{skill} phase. The task is considered successful if this to-go distance is less than 1 cm. \\
    \textbf{Real world.} We employ a 7-DoF Franka Panda equipped with a wrist-mounted Intel RealSense D435 RGB-D camera. Note that since our method is RGB only, the depth images are only used for another baseline. For all the grasping tasks, we define success when the robot manages to grasp and lift the target object for at least 5s. For the trash into bin task, we define success when the trash paper is placed in the bin. For the opening drawer of air fryer task, we define success when the drawer of air fryer is opened by the gripper. \\
    \textbf{Computing Statistics.} We use Nvidia RTX 4090 GPU for data augmentation, policy training and inference. The system runtime for OP-Gen is as follows: a) Data Collection: $\sim10s$; b) 3D Generation: $\sim120s$, including $\sim100s$ for EscherNet rendering (100 views), and $\sim20s$ for Instant-NGP training; c) Data Augmentation (20k augmented points): $\sim328s$, including  $\sim300s$ data points generation during Anchored Trajectory Generation, and $\sim28s$ Instant-NGP rendering; d) Policy training: $\sim138 \space mins$.
    \vspace{-0.1cm}
    \subsection{Can OP-Gen solve a range of everyday tasks from a single demonstration and how does it compare to baselines?}
    We evaluate OP-Gen on six real-world tasks, conducting 20 rollouts per task using four different baselines. During each rollout, the EEF starts at different poses sampled from the working space. Among these, 10 initial poses are sampled within a ±45$^\circ$ fan-shaped range around the object, with the central axis aligned to the first demo camera view. These poses are categorised as \textbf{Narrow} initial poses, while the remaining poses sampled outside this range are categorised as \textbf{Omni} initial poses. We compare our OP-Gen with baselines including: 
    (1) \textbf{No Aug}: single demonstration without augmentation. This represents the lower bound of the method. (2) \textbf{GGMP}: generated-geometry-based motion plan. At test time, we first estimate the camera pose relative to the bottleneck pose using VGGT \cite{wang2025vggt}. Based on this estimated pose and the generated object geometry, we then run a motion planner CuRobo to generate a trajectory and execute it. (3) \textbf{OP-PCD}: we reconstruct a partial point cloud of the target object using RGB-D images from the single demonstration. For best practice, we remove points that are further away from their neighbours on average. Then we project point clouds onto 2D images rather than rendering from our 3D generation module. Comparison with this baseline will show the benefits of using complete 3D generation rather than partial reconstruction. (4) \textbf{SPARTN}: we use our implementation of SPARTN \cite{zhou2023nerf}, which builds a partial NeRF using camera views along the demonstration trajectory. Then we augment the dataset as our main method. This method can provide plausible data augmentation near the demonstration views but no additional information from the unseen part of the object. (5) \textbf{Upper Bound} (UB): full NeRF based augmentation. We replace our 3D generation module with the previous full-scan NeRF of the object. Note that since we're investigating one-shot imitation learning, demonstration views will be inadequate for the full NeRF training. As a result, we only consider this method as an upper bound rather than a fair baseline. To highlight the data efficiency of our method, we also record the data collection time for each of the baseline methods.

    \begin{table*}[!htbp]
    \scriptsize
    \centering
    \caption{Success rates for OP-Gen and 5 baselines across 6 real-world tasks (\%), as well as the corresponding data collection time (s). 10 rollouts each in both Narrow and Omni settings. Data statistics are presented using 95\% Wilson Confidence Intervals (WCI), which is shown in parentheses after the total averages.}
    \label{tab:tabel-1}
    \begin{tabular}{l|c@{\hskip10.5pt}c@{\hskip10.5pt}c@{\hskip10.5pt}c@{\hskip10.5pt}c@{\hskip10.5pt}c@{\hskip10.5pt}c@{\hskip10.5pt}c@{\hskip10.5pt}c@{\hskip10.5pt}c@{\hskip10.5pt}c@{\hskip10.5pt}c@{\hskip10.5pt}c@{\hskip10.5pt}cccc}
        \toprule
        Level & \multicolumn{7}{c}{Narrow} & \multicolumn{7}{c}{Omni} & \multirow{2.5}{*}{\makecell{Total \\ Average}} & \multirow{2.5}{*}{WCI (95\%)} & \multirow{2.5}{*}{Time(s)}  \\
        \cmidrule(r){0-0} \cmidrule(lr){2-8} \cmidrule(lr){9-15} 
        Method & \textit{Drill} & \textit{Mug} & \textit{Plane} & \textit{Coffee} & \textit{Bin} & \textit{Fryer} & \textit{Avg} & \textit{Drill} & \textit{Mug} & \textit{Plane} & \textit{Coffee} & \textit{Bin} & \textit{Fryer} & \textit{Avg} &  &   \\
        \midrule
        No Aug & 0 & 0 & 0 & 0 & 0 & 0 & 0.0 & 0 & 0 & 0 & 0 & 0 & 0 & 0.0 & 0.00 & (0.00, 3.10) & 10\\
        GGMP & 60 & 0 & 0 & 20 & 0 & 10 & 15.0 & 70 & 10 & 0 & 30 & 0 & 0 & 18.3 & 16.70 & (11.06, 24.35) & 10 \\
        OP-PCD & 0 & 0 & 30 & 20 & 40 & 20 & 18.3 & 0 & 0 & 0 & 0 & 20 & 10 & 5.0 & 11.65 & (7.08, 18.63) & 10 \\
        SPARTN  & 20 & 20 & 30 & 30 & 60 & 10 & 28.3 & 0 & 20 & 0 & 0 & 0 & 0 & 3.3 & 15.80 & (10.38, 23.41) & 10 \\
        \rowcolor{blue!10} 
        OP-Gen & \textbf{90} & \textbf{90} & \textbf{70} & \textbf{90} & \textbf{80} & \textbf{90} & \textbf{85.0} & \textbf{90} & \textbf{60} & \textbf{80} & \textbf{60} & \textbf{80} & \textbf{70} & \textbf{73.3} & \textbf{79.15} & \textbf{(71.05, 85.47)} & 10 \\
        \midrule
        UB & 90 & 90 & 80 & 80 & 80 & 90 & 85.0 & 100 & 80 & 80 & 80 & 80 & 80 & 83.3 & 84.15 & (76.59, 89.62) & 90 \\
        \bottomrule
    \end{tabular}
    \vspace{-0.4cm}
\end{table*}

    All real-world experiment results are summarised in Table~\ref{tab:tabel-1}. Across all methods, performance in the \textbf{Narrow} setting consistently surpasses that in the \textbf{Omni} setting. Without any data augmentation, policies fail during all rollouts due to out-of-distribution states, confirming that a single demonstration is insufficient for effective behavioural cloning. For GGMP, while VGGT can estimate relative camera poses, we found that the results often lack scale information, leading to incorrect movements and occasional collisions. This issue arises because RGB-only pose estimation methods cannot reliably recover absolute scale. Consequently, accurate and scaled pose estimation becomes the key bottleneck for such planning-based methods. Augmentation via OP-PCD slightly improves performance, but due to a significant rendering gap as shown in Fig. \ref{fig:rendering comp}, the resulting policy handles fewer than half of the initial poses, achieving only an 18.3\% average success rate in the \textbf{Narrow} setting and 5\% in the \textbf{Omni} setting. SPARTN's partial NeRF reconstruction yields a 28.3\% success rate in the \textbf{Narrow} setting but still fails in most \textbf{Omni} cases, due to missing information from novel viewpoints. In contrast, our proposed method achieves an average success rate of 85\% from the \textbf{Narrow} regions and 73.3\% from the \textbf{Omni} regions, significantly outperforming all baselines and approaching the performance of the upper bound. Notably, our method maintains similar performance across both settings, demonstrating that 3D generative models can provide informative representations even under drastic viewpoint changes. During testing, we observed that nearly all failure cases stemmed from last-inch inaccuracies—either due to premature gripper command or collisions. We attribute these errors to subtle misalignments between the generated 3D model and the physical object, as well as a limited number of grasp action labels, leading to a multi-modal distribution in the final EEF action predictions. Overall, these results demonstrate that 3D generative models serve as effective tools for data augmentation, enhancing the robustness of behavioural cloning policies. Additionally, they contribute valuable information beyond the original demonstration distribution, substantially reducing the need for large amounts of real-world training data.

    \subsection{How critical is 3D generation quality for OP-Gen?}

\begin{figure}[htbp]
    \centering
    \subfigure{\includegraphics[width=0.18\linewidth]{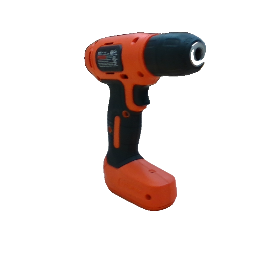}}
    \subfigure{\includegraphics[width=0.18\linewidth]{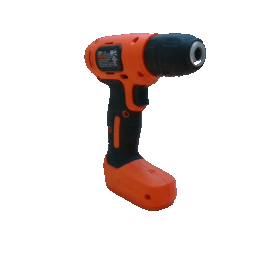}}
    \subfigure{\includegraphics[width=0.18\linewidth]{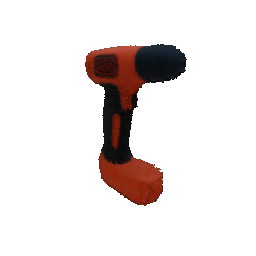}}
    \subfigure{\includegraphics[width=0.18\linewidth]{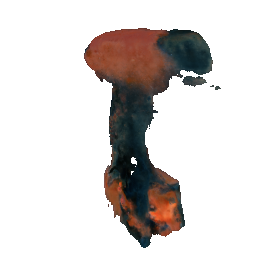}}
    \subfigure{\includegraphics[width=0.18\linewidth]{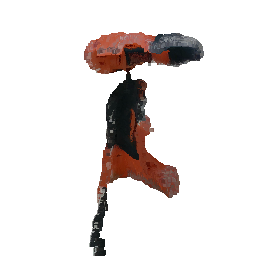}}
    \setcounter{subfigure}{0}\\\vspace{-0.5cm}
    \subfigure{\includegraphics[width=0.18\linewidth]{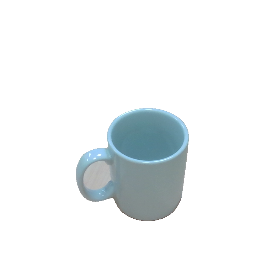}}
    \subfigure{\includegraphics[width=0.18\linewidth]{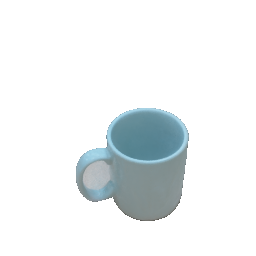}}
    \subfigure{\includegraphics[width=0.18\linewidth]{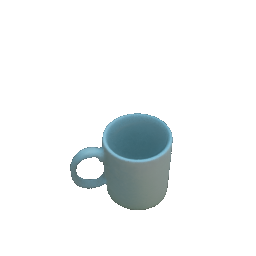}}
    \subfigure{\includegraphics[width=0.18\linewidth]{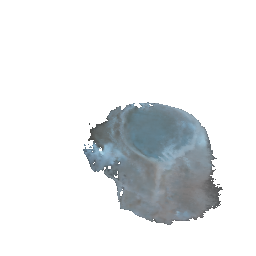}}
    \subfigure{\includegraphics[width=0.18\linewidth]{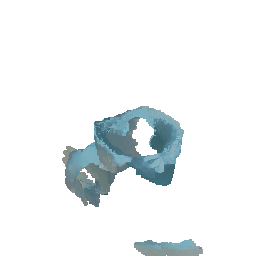}}
    \setcounter{subfigure}{0}\\\vspace{-0.5cm}
    \subfigure{\includegraphics[width=0.18\linewidth]{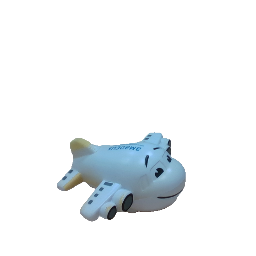}}
    \subfigure{\includegraphics[width=0.18\linewidth]{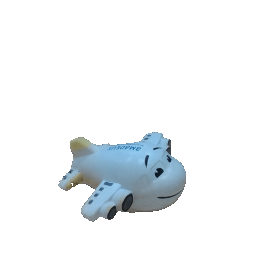}}
    \subfigure{\includegraphics[width=0.18\linewidth]{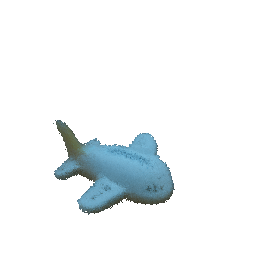}}
    \subfigure{\includegraphics[width=0.18\linewidth]{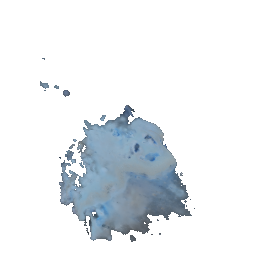}}
    \subfigure{\includegraphics[width=0.18\linewidth]{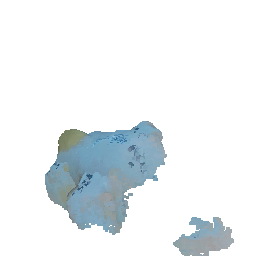}}
    \setcounter{subfigure}{0}\\\vspace{-0.5cm}
    \subfigure{\includegraphics[width=0.18\linewidth]{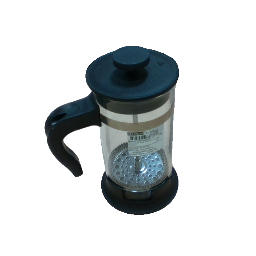}}
    \subfigure{\includegraphics[width=0.18\linewidth]{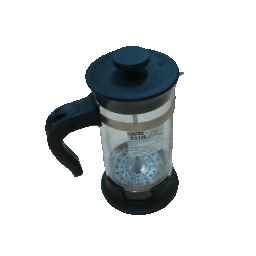}}
    \subfigure{\includegraphics[width=0.18\linewidth]{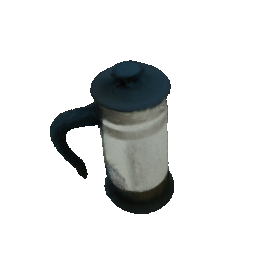}}
    \subfigure{\includegraphics[width=0.18\linewidth]{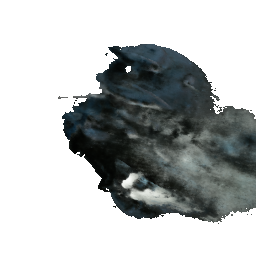}}
    \subfigure{\includegraphics[width=0.18\linewidth]{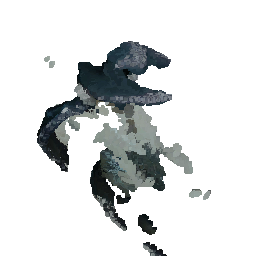}}
    \setcounter{subfigure}{0}\\\vspace{-0.3cm}
    \subfigure{\includegraphics[width=0.18\linewidth]{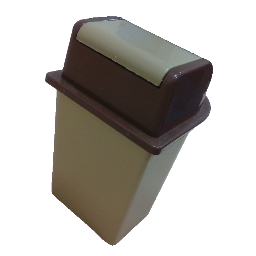}}
    \subfigure{\includegraphics[width=0.18\linewidth]{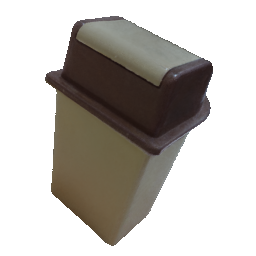}}
    \subfigure{\includegraphics[width=0.18\linewidth]{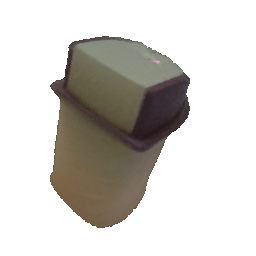}}
    \subfigure{\includegraphics[width=0.18\linewidth]{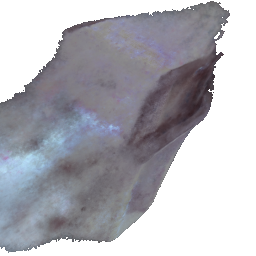}}
    \subfigure{\includegraphics[width=0.18\linewidth]{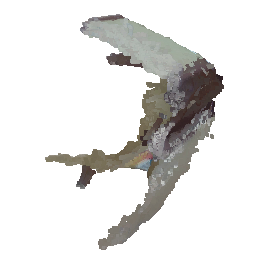}}
    \setcounter{subfigure}{0}\\\vspace{-0.3cm}
    \subfigure[GT]{\includegraphics[width=0.18\linewidth]{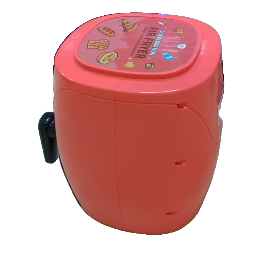}}
    \subfigure[UB]{\includegraphics[width=0.18\linewidth]{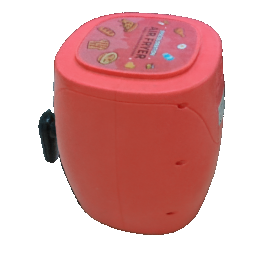}}
    \subfigure[OP-Gen]{\includegraphics[width=0.18\linewidth]{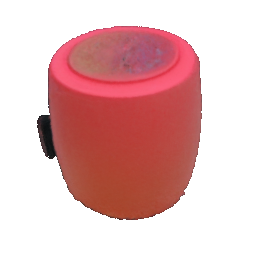}}
    \subfigure[SPARTN]{\includegraphics[width=0.18\linewidth]{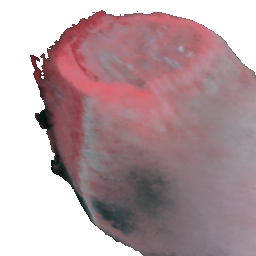}}
    \subfigure[OP-PCD]{\includegraphics[width=0.18\linewidth]{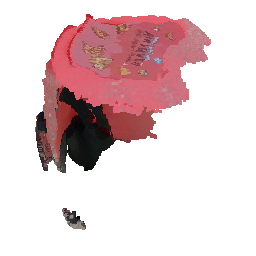}}
    \caption{Qualitative results showing renders from different novel views across all methods.}    \label{fig:rendering comp}
    \vspace{-10pt}
\end{figure}

    To evaluate the impact of the 3D generation model on policy performance, we present qualitative comparisons of novel view renderings produced by different baselines in Fig. \ref{fig:rendering comp}. As shown, the UB generates renderings that are nearly identical to the ground truth (GT), thereby supporting strong omnidirectional policy performance. Our proposed OP-Gen also produces renderings that are visually similar to the GT, maintaining sufficient fidelity for effective policy training. In contrast, renderings from SPARTN and OP-PCD deviate significantly from the GT, lacking even structural correspondence when changing to the opposite of the input views, which hampers the methods' ability to learn OP effectively from these views.

\begin{figure} [htbp]
    \centering
    \includegraphics[width=0.9\linewidth]{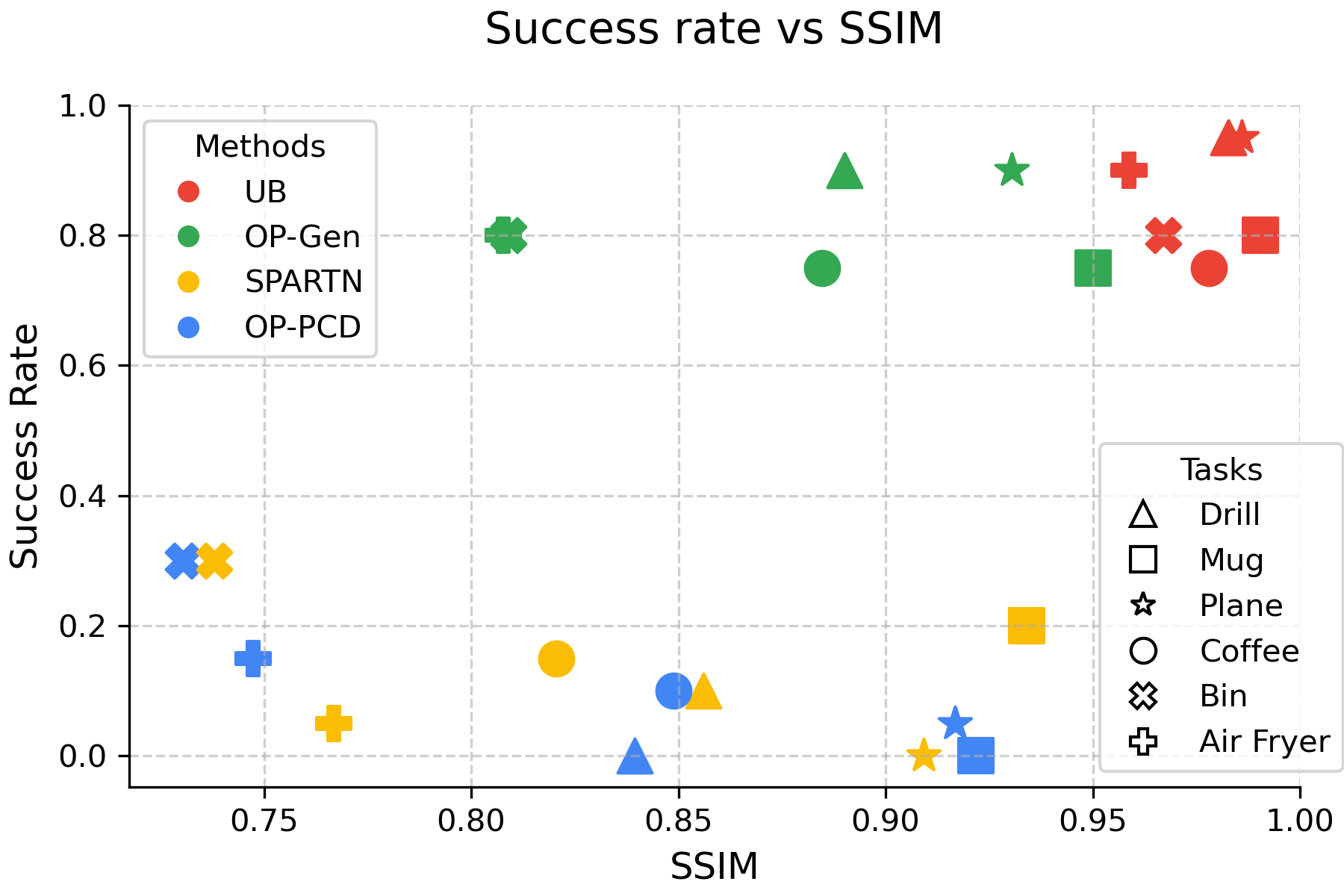}
        \caption{Relationship between task success rate and SSIM value across different tasks and baselines.}
        \label{fig:SSIM}
        \vspace{-12pt}
    \end{figure}

    To quantitatively evaluate the impact of 3D generation and reconstruction quality, we measure the average SSIM \cite{wang2004image} scores between renderings and GT images along the full NeRF scan trajectory across all methods, reflecting their structural similarity. Fig.~\ref{fig:SSIM} illustrates how SSIM affects the success rate. As shown, full-scan UB consistently achieves the highest SSIM and success rates, establishing it as the upper bound for OP learning. Our OP-Gen method, utilising EscherNet \cite{kong2024eschernet}, attains comparable success within the SSIM range of 0.8 – 0.95. However, SPARTN and OP-PCD achieve less than 30\% success despite SSIM values ranging from 0.7 to 0.95. Although these ranges overlap, most SSIM values for SPARTN and OP-PCD cluster between 0.7 and 0.85, with only the \textit{mug} and \textit{plane} tasks reaching higher SSIM values. This exception is due to the textureless surfaces of these objects, which simplify reconstruction. Meanwhile, OP-Gen consistently achieves SSIM 
    between 0.85 and 0.95 across tasks, leaving only two large complex objects \textit{bin} and \textit{air fryer} out of range, reflecting the increased difficulty of accurate reconstruction. Based on these observations, we conclude that an SSIM above 0.95 is necessary for small, textureless objects to achieve high policy performance, while an SSIM above 0.8 suffices for larger, more complex objects. 
    \vspace{-10pt}
    
    Across tasks, we also observe a correlation between higher SSIM scores and increased policy success rates. Interestingly, when SSIM falls below 0.95, it becomes important to examine the consistency between views. For instance, OP-PCD and SPARTN achieve SSIM around 0.95 near the input angles but drop below 0.5 when rendering from opposite directions, making object structure unrecognizable as in Fig. \ref{fig:rendering comp}. In contrast, OP-Gen maintains consistent SSIM around 0.85 across all viewpoints, enabling reliable object identification. Therefore, while higher image fidelity improves policy performance, consistency across viewpoints is even more crucial for omnidirectional policy learning.

    \subsection{How would different trajectory generation pipelines affect the performance?}
    To assess each component’s impact in the anchored trajectory generation pipeline, we conduct an ablation study across six tasks. We systematically remove components or introduce variations and report the average to-go distance to the bottleneck in simulation, as well as average success rates in both simulation and real-world settings. As shown in Table~\ref{tab:tabel-2}, we ablate four implementations. First, removing the view re-focus (w/o VRF) module significantly reduces performance, as objects may leave the camera's view, leading to ineffective datasets. Second, using only motion planning (MP only), without VRF or SLERP interpolation, yields a 46.67\% real-world success rate, suggesting SLERP adds subtle view-focusing constraints. Third, applying VRF at all motion-planned trajectory points without SLERP (MP + VRF) drops the success rate to 4.17\% due to abrupt rotations, especially near the trajectory center, resulting in unstable action labels. Lastly, a linear variant (Linear) without collision-free planning achieves a 43.33\% real-world success rate, with most failures attributed to collisions. Since the simulator lacks collision checking, this variant shows artificially lower to-go distances and inflated simulated success.
    \vspace{-6pt}
    \begin{table}[!hbtp]
        \centering
        \caption{Ablation study on different building blocks for anchored trajectory generation.}
        \label{tab:tabel-2}
        \begin{tabular}{lccc}
            \toprule
            Method & \textit{Simulation Dist} & \textit{Simulation SR} & \textit{Real World SR} \\
            \midrule
            OP & 0.013 & 83.33 & 79.17 \\
            w/o VRF & 0.082 & 80.83 & 53.33 \\
            MP only & 0.152 & 75 & 46.67 \\
            MP + VRF & 0.123 & 6.67 & 4.17 \\
            Linear & 0.009 & 95 & 43.33 \\
            \bottomrule
        \end{tabular}
        \vspace{-0.3cm}
    \end{table}
    \vspace{-6pt}

%===============================================================================

\section{Conclusion} 
\label{sec:conclusion}
We present OP-Gen, a method that transforms a single wrist-mounted camera demonstration into a large omnidirectional training set using object-centric 3D generative models. Through real-world experiments, we show that the resulting policies generalise to multiple unseen initial states and outperform point cloud and partial NeRF based augmentation baselines, approaching an upper bound obtained with full object scans. We further show that although higher-quality novel-view renderings improve performance, cross-view consistency is more critical for omnidirectional policy learning. Ablations demonstrate that the anchored-trajectory pipeline is essential, as removing it substantially degrades real-world success. Overall, 3D generative models provide a practical route to 3D augmentation for one-shot imitation learning, and continued advances in 3D generation may further strengthen omnidirectional policy robustness and generalisation.

\section{Limitations} 
\label{sec:limitations}
Our approach targets a constrained but practical setting: single-object, single-stage manipulation where the target remains within the wrist-mounted RGB camera’s field of view. Because OP-Gen relies on object-centric reconstruction and rendering, it does not handle persistent occlusions, multi-object interactions, or cases where a grasped object blocks a subsequent target. Addressing such scenarios would require occlusion-aware, multi-object reconstruction, which we leave for future work.

The framework is also limited to single-stage interactions. Extending OP-Gen to long-horizon, multi-stage tasks with multiple objects would require additional mechanisms for object tracking, stage transitions, and partial observability handling, which are beyond the scope of this work.

OP-Gen further depends on the fidelity of the generated 3D object models. While moderate geometric inaccuracies can be tolerated in our tasks, reconstruction errors may degrade fine manipulation requiring precise contact. We analyse this sensitivity through experiments, but a broader characterisation of failure modes remains future work.

Finally, the proposed method is instance-specific, as policies are trained on augmented observations of a single demonstrated object. Although 2D augmentation may improve appearance robustness, OP-Gen does not directly address category-level generalisation to unseen objects. Integrating object-centric augmentation with broader representation learning is a promising direction for future research.

\bibliographystyle{IEEEtran}
\bibliography{example}

\end{document}